\pdfoutput=1

\documentclass[11pt]{article}

\usepackage[final]{acl}

\usepackage{times}
\usepackage{latexsym}

\usepackage[T1]{fontenc}

\usepackage[utf8]{inputenc}

\usepackage{microtype}

\usepackage{inconsolata}

\usepackage{graphicx}
\usepackage{subcaption}
\usepackage{amsmath}
\usepackage{amssymb}
\usepackage{multirow}
\usepackage{adjustbox}
\usepackage{algorithm}
\usepackage{algpseudocode}
\usepackage{pgffor}
\algrenewcommand\algorithmicrequire{\textbf{Input:}}
\algrenewcommand\algorithmicensure{\textbf{Output:}}
\usepackage[labelformat=simple]{subcaption}
\usepackage{hyperref}

\title{Eigen  Attention: Attention in Low-Rank Space for KV Cache Compression}

\author{Utkarsh Saxena, \hspace{4pt}   Gobinda Saha,  
 \hspace{4pt} Sakshi Choudhary, 
  \hspace{4pt} Kaushik Roy \\
  Purdue University \\
  \texttt{\{saxenau, gsaha, choudh23, kaushik\}@purdue.edu} 
  }

\begin{document}
\maketitle
\begin{abstract}
  

  Large language models (LLMs) represent a groundbreaking advancement in the domain of natural language processing due to their impressive reasoning abilities. Recently, there has been considerable interest in increasing the context lengths for these models to enhance their applicability to complex tasks. However, at long context lengths and large batch sizes, the key-value (KV) cache, which stores the attention keys and values, emerges as the new bottleneck in memory usage during inference. To address this, we propose Eigen Attention, which performs the attention operation in a low-rank space, thereby reducing the KV cache memory overhead. Our proposed approach is orthogonal to existing KV cache compression techniques and can be used synergistically with them. Through extensive experiments over OPT, MPT, and Llama model families, we demonstrate that Eigen Attention results in up to 40\% reduction in KV cache sizes and up to 60\% reduction in attention operation latency with minimal drop in performance. Code is available at \href{https://github.com/UtkarshSaxena1/EigenAttn/tree/main}{https://github.com/UtkarshSaxena1/EigenAttn}.
  
\end{abstract}

\section{Introduction}

The recent boom in artificial intelligence applications and their widespread public adoption can be attributed to the human-like capabilities of large language models (LLMs).~LLMs have demonstrated remarkable performance across a wide range of natural language processing (NLP) tasks \cite{llm-leaderboard}. The maximum number of tokens these models can process simultaneously to understand and generate text is referred to as the context/sequence length, and this closely determines their performance limits. Hence, there has been considerable interest in increasing the context length to enhance their capabilities \cite{longcontext-arxiv2024, longrope-arxiv2024, gpt4-arxiv2023}. Longer context lengths open up new possibilities, such as summarizing lengthy documents, retrieving information to answer questions about extensive texts, and analyzing code. To make applications enabled by LLMs more accessible, developing techniques to serve these models efficiently is crucial. One standard technique to accelerate LLM inference on GPUs is caching the intermediate attention keys and values through a KV cache to avoid expensive re-computation for every generated token. However, it is observed that at long context lengths, the KV cache becomes the new memory and latency bottleneck \cite{pope2022efficiently}. Furthermore, while batching multiple requests together amortizes the weight access cost of LLMs, it exacerbates the KV cache memory overhead. Consider the weight and KV cache memory footprint for the Llama-2 7B model with a batch size of 16 and a context length of 32k tokens. Here, the weight memory occupies 14 GB, while the KV cache requires a significantly higher memory of 256 GB at 16-bit precision. In essence, increasing the context/sequence length of LLMs has transformed LLM inference into a memory-bound problem. The entire KV cache must be bought on-chip from the off-chip GPU memory for each newly generated token while the computation core stays idle, waiting for data. 
\begin{figure*}[ht!]
    \centering
    \includegraphics[width=\textwidth]{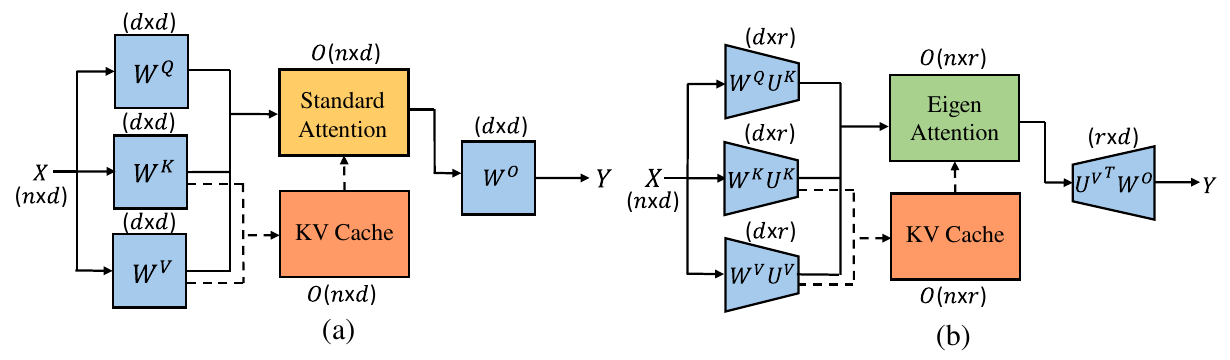}
    \caption{Comparison between (a) Standard Attention and (b) Eigen Attention. Eigen Attention utilizes lower dimensional ($r\ll d$) query, key, and value projection matrices than the standard attention operation, leading to KV cache compression and compute FLOPs benefits.} \label{fig:eigen_attn_overview}
\end{figure*}
\begin{table*}[ht]
\caption{KV cache size, number of parameters, and total FLOPs for Standard vs Eigen Attention computed for sequence length $n$, hidden dimension $d$ in standard attention and hidden dimension $r$ ($\ll d$) in Eigen Attention. }
\vspace{-3mm}
\label{tab:eigenmetrics}
\small
\begin{center}
\begin{tabular}{ccc}
\hline
 & Standard Attention & Eigen Attention\\
\hline
\hline
KV Cache Size & $2.n.d$ & $2.n.r$\\
\hline
\# Parameters & $4.d^2$ & $4.d.r$\\
\hline
FLOPs (generation phase) & $4.d^2 + 2.n.d$ & $4.d.r+2.n.r$\\
 \hline
\end{tabular}
\end{center}
\end{table*}

Existing methods to address the KV cache bottleneck can be broadly classified into four distinct categories. First, some approaches focus on reducing the number of KV cache heads in the multi-head attention block through grouped query and multi-query attention \cite{groupedqueryattention-arxiv2023,multiqueryattention-arxiv2019}. Second, a few methods aim to alleviate the memory overhead by utilizing a low-precision quantized KV cache \cite{kvquant-arxiv2024,kivi-arxiv2024}. Third, certain strategies involve evicting KV cache values associated with unimportant tokens, thereby caching only the keys and values of important tokens determined through some metrics \cite{h20-neurips2023,keyformer-mlsys2024}. Finally, some approaches tackle the issue from a systems perspective by utilizing the CPU and disk memory for the KV cache or by extending virtual memory and incorporating paging techniques into the attention mechanism  \cite{pagedattention-arxiv2023}.

In contrast to the above-mentioned techniques, we propose Eigen Attention, a strategy to alleviate the KV cache overhead through low-rank approximation. Eigen Attention leverages the observation that attention inputs in LLMs (i.e., key, query, and value) can be reasonably approximated using a few principal basis vectors or eigenvectors \cite{Yu_Wu_2023}. Expressing keys and values as a linear combination of these principal vectors essentially reduces their dimension, leading to a lower memory footprint of the KV cache. To achieve this, we use a very small subset of training data as a calibration dataset to generate a set of query, key, and value matrices for the trained model. Subsequently, we obtain the basis vectors through Singular Value Decomposition (SVD) on these matrices and choose the most important directions through a pre-defined error threshold.
Eigen Attention is a post-training technique that can be applied without requiring any additional fine-tuning. Moreover, our approach is orthogonal to existing techniques for KV cache compression and can be used in conjunction with them. Figure \ref{fig:eigen_attn_overview} illustrates the differences between standard attention and our proposed Eigen Attention. Columns of the $\textbf{U}^K$ matrix are the principal basis vectors for keys and queries. Similarly, $\textbf{U}^V$ contains the important basis vectors for the value matrix. We project the attention inputs into a low-rank space defined by the eigenvectors $\textbf{U}^K$ and $\textbf{U}^V$ by assimilating these projections into the corresponding weight matrices $\mathbf{W}^Q$, $\mathbf{W}^K$, $\mathbf{W}^V$ and $\mathbf{W}^O$ offline. During inference, the low-dimensional $\mathbf{W}^K\mathbf{U}^K$ and $\mathbf{W}^V\mathbf{U}^V$ are multiplied with the input vector $\mathbf{X}$ to generate lower-dimensional K and V matrices. While the primary goal of Eigen Attention is to reduce the memory overhead of the KV cache, the proposed low-rank approximation also leads to a compute-efficient implementation of the attention block (Section \ref{sec:results}). For the standard dimension $d$ and the compressed dimension $r (\ll d)$, Table \ref{tab:eigenmetrics} shows the reduction in KV cache size and floating point operations (FLOPs) achieved through Eigen Attention.
In summary, we make the following contributions:
\begin{itemize}
    \item We propose Eigen Attention, a novel mechanism to efficiently serve LLMs by compressing the KV cache through low-rank approximations.
    \item We demonstrate that the low-rank approximation employed by Eigen Attention enhances the compute efficiency of the attention block.
    \item Extensive experiments across various models and language tasks show that Eigen Attention compresses the KV cache by up to 40\% along with upto 60\% reduction in the attention operation latency.
\end{itemize}



\section{Background}
\textbf{Multi-Head Attention.}
A typical LLM consists of $L$ decoder layers, each with two components: multi-head attention (MHA) and the fully connected feed-forward network (FFN). For an input token embedding \( \mathbf{X} \in \mathbb{R}^{n \times d} \), the MHA block performs attention operations in parallel across \(h\) heads. Here, \(n\) is the sequence length, and \(d\) is the hidden dimensionality of the model. For each attention head \(i \in \{1,2,...h\}\), \(\mathbf{X}\) is transformed into key, query, and value matrices as follows:
\begin{equation}\label{eq:qkv}
\mathbf{Q}_i = \mathbf{X}\mathbf{W}_i^Q, \hspace{2mm} \mathbf{K}_i = \mathbf{X}\mathbf{W}_i^K, \hspace{2mm} \mathbf{V}_i = \mathbf{X}\mathbf{W}_i^V
\end{equation}
Here, \( \mathbf{W}_i^Q \in \mathbb{R}^{d \times d_h}, \mathbf{W}_i^K  \in \mathbb{R}^{d \times d_h}, \mathbf{W}_i^V  \in \mathbb{R}^{d \times d_h} \) are learnable weight matrices with \(d_h\) = \(\frac{d}{h}\). The attention operation $\mathbf{A}$ at each head \(i\) is computed, and the results from all heads are concatenated to obtain the final output of the MHA block, as shown in Equation \ref{eq:attn}.
\vspace{-2mm}
\begin{equation}
\label{eq:attn}
\begin{split}
&\text{h}_i=\mathbf{A}(\mathbf{Q}_i, \mathbf{K}_i, \mathbf{V}_i)=\text{Softmax}\left(\frac{\mathbf{Q}_i \mathbf{K}_i^T}{\sqrt{d_h}}\right) \mathbf{V}_i,\\
&\text{MHA}(\mathbf{Q}, \mathbf{K}, \mathbf{V}) = [\text{h}_1, \text{h}_2, .., \text{h}_h]\mathbf{W}^O
\end{split}
\end{equation}

\textbf{LLM Inference.}\label{sec:infer}
The inference consists of the prefill and generation phases. In the prefill phase, the keys $\mathbf{K}_i$ and $\mathbf{V}_i$ values are computed for an input token embedding \( \mathbf{X} \in \mathbb{R}^{b \times n \times d} \) with batch size $b$ (as shown in Equation \ref{eq:qkv}) and cached in memory for the generation phase. The total size of KV cache (in bits) can be derived by $2 * b* n * d * h* L * p$, where $L$ corresponds to the number of decoder layers in the LLM, and $p$ corresponds to the precision of cached vectors. 

In the generation phase, the model uses and updates the KV cache to generate the output auto-regressively, one token at a time. Note that this phase is memory-bound. For an incoming token embedding $\mathbf{x} \in \mathbb{R}^{b \times 1 \times d}$, the key $\mathbf{k}_i$ and value $\mathbf{v}_i$ computed through Equation \ref{eq:qkv} are appended to the KV cache:
    \[ \mathbf{K}_i \leftarrow \text{Concat}(\mathbf{K}_i, \mathbf{k}_i),\mathbf{V}_i \leftarrow \text{Concat}(\mathbf{V}_i, \mathbf{v}_i) \]

The query $\mathbf{q}_i$ obtained through Equation \ref{eq:qkv} is used to compute the attention output for each head:
\begin{equation}
        \text{h}_i= \text{Softmax}\left(\mathbf{q}_i \mathbf{K}_i^T/\sqrt{d_h}\right) \mathbf{V}_i
\end{equation}
Then, similar to Equation \ref{eq:attn}, the heads' outputs are concatenated and multiplied with $\mathbf{W}^O$ to produce the final output of the attention block.

\section{Related Works}\label{sec:related_works}
\vspace{-2.25mm}
\textbf{KV Cache Compression.}
As mentioned in Section \ref{sec:infer}, the KV cache size is computed as $2 * b* n * d * h* L * p$. KV cache compression methods target these factors to reduce its overall memory footprint. Multi-query attention \cite{multiqueryattention-arxiv2019} and grouped query attention \cite{groupedqueryattention-arxiv2023} reduce the number of attention heads $h$. Quantization based methods reduce the precision $p$ \cite{mikv-arxiv2024, gear-arxiv2024, kivi-arxiv2024, kvquant-arxiv2024}. Notably, works like KIVI \cite{kivi-arxiv2024} and KV Quant \cite{kvquant-arxiv2024} have demonstrated that the precision of $\mathbf{K}$ and $\mathbf{V}$ matrices can be reduced to as low as 2-bits. Several works attempt to reduce the sequence length $n$ by only caching $\mathbf{K}$ and $\mathbf{V}$ corresponding to a subset of tokens \cite{longformer}. Another strategy is to cache K and V according to token importance \cite{h20-neurips2023, keyformer-mlsys2024, scissorhands-neurips2024, atp-arxiv2024, snapKV-arxiv2024, pyramidkv-arxiv-2024}. $\textnormal{H}_2\textnormal{O}$ \cite{h20-neurips2023} finds important tokens by monitoring accumulated attention scores. KeyFormer \cite{keyformer-mlsys2024} improves $\textnormal{H}_2\textnormal{O}$ by considering the importance of discarded tokens as well. Recently, \cite{layercondensed-arxiv2024} demonstrated that the cached $\mathbf{K}$ and $\mathbf{V}$ can be reused across the decoder layers, essentially reducing $L$. In contrast to these techniques, Eigen Attention reduces the embedding dimension $d$ of each cached $\mathbf{K}$ and $\mathbf{V}$ vector. It is orthogonal to the existing KV cache compression techniques and can be used in conjunction with them. 
\begin{figure*}[h!]
    \begin{subfigure}[t]{0.24\textwidth}
        \includegraphics[height=1.6in]{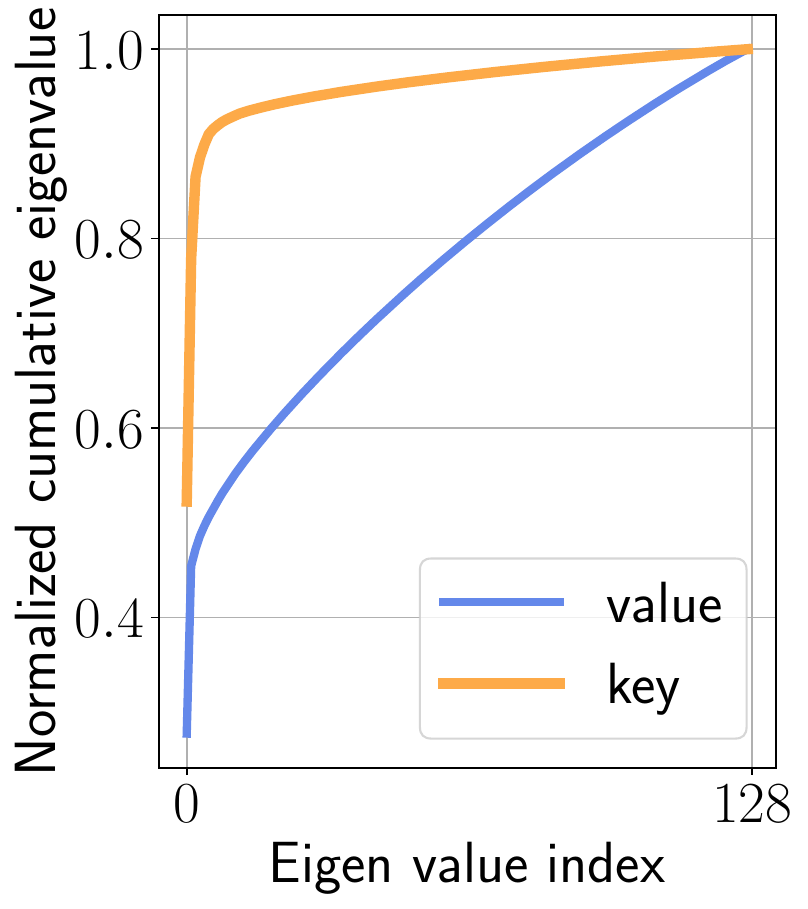}
        \caption{Layer \#15 Head \#16}
    \end{subfigure}%
   \begin{subfigure}[t]{0.24\textwidth}
        \includegraphics[height=1.6in]{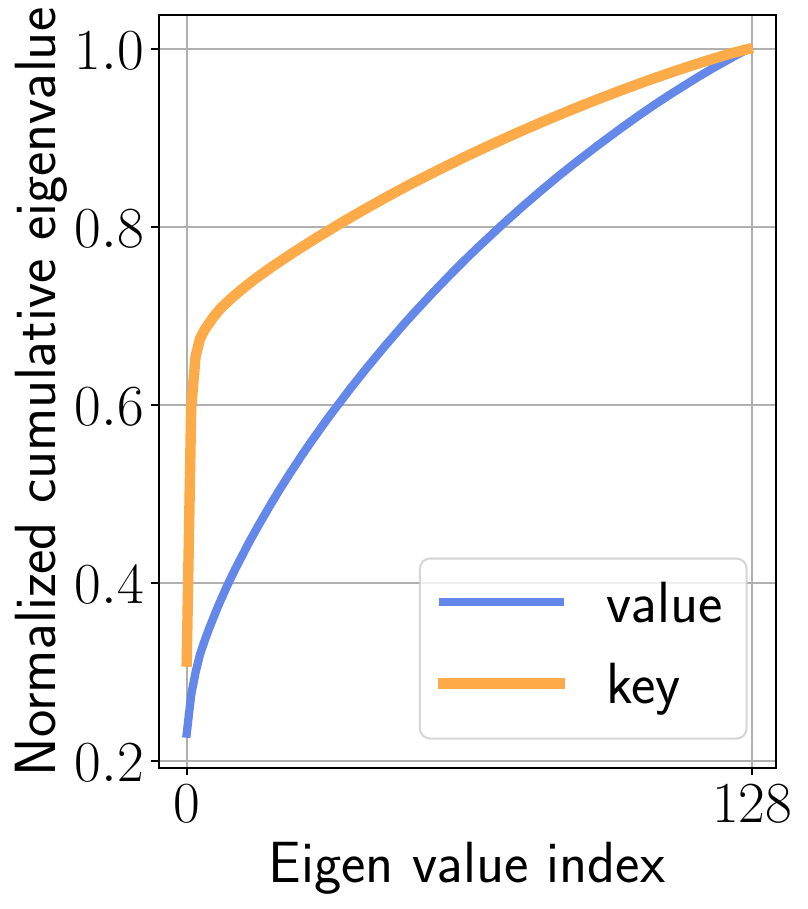}
        \caption{Layer \#25 Head \#16}
    \end{subfigure}%
    \begin{subfigure}[t]{0.24\textwidth}
        \includegraphics[height=1.6in]{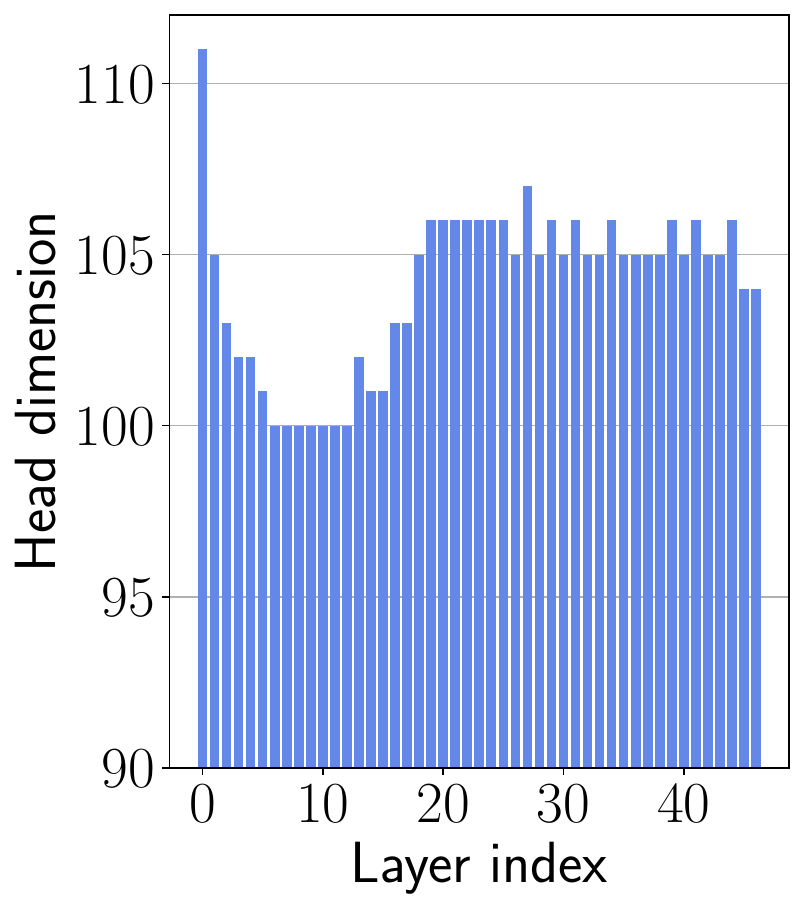}
        \caption{Value}
    \end{subfigure}%
    \begin{subfigure}[t]{0.24\textwidth}
        \includegraphics[height=1.6in]{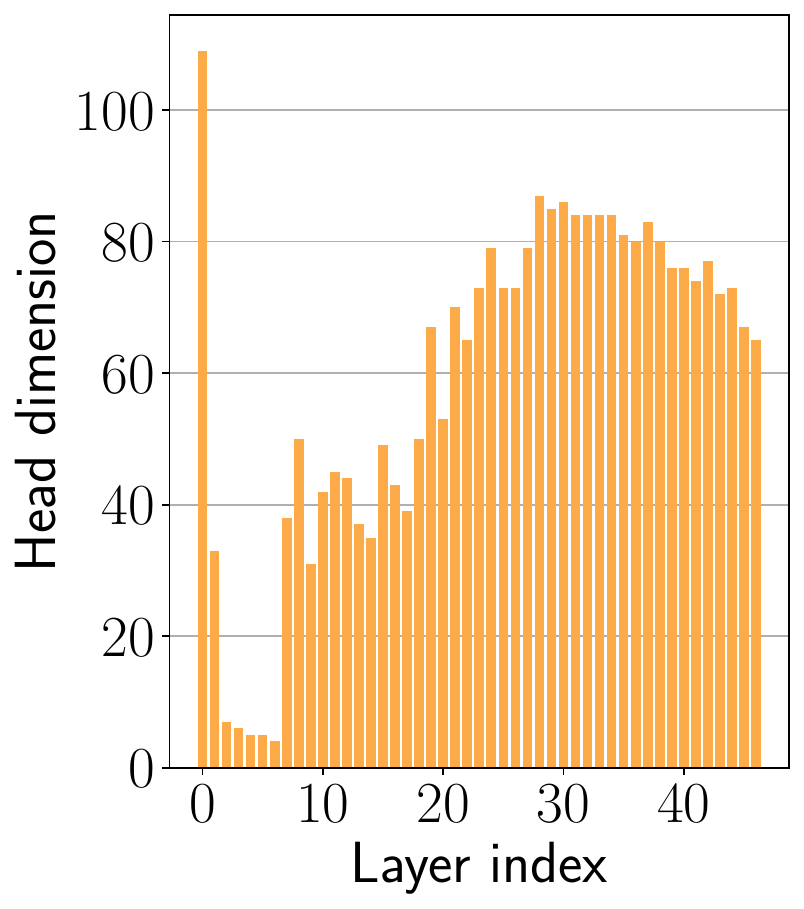}
        \caption{Key}
    \end{subfigure}%
    \caption{Eigenvalue spectrum analysis for OPT-30b model. (a), (b) The Y-axis is the normalized cumulative eigenvalue value after performing SVD on the key value representation matrix, and the X-axis is an index of the largest eigenvalue. (c), (d) Dimensions of the low-rank matrices with normalized cumulative eigenvalue of 0.9.}\label{fig:low_rankness}
\end{figure*}

\textbf{Low-Rank Approximation.} 
Various works in literature have leveraged low-rank approximation for performing efficient LLM inference. Recent works \cite{Yu_Wu_2023, rankdiminish} have shown that while the weight matrices for transformers-based models are not inherently sparse, the activations are. LoRD \cite{lord} leverages this observation to compress the weight matrix of LLMs by representing it as a product of two low-rank matrices. LoSparse \cite{losparse} combines pruning and low-rank approximation and expresses the weight matrix as a sum of a low-rank matrix and a sparse matrix. Fisher-weighted SVD \cite{fwsvd} proposes to utilize Fisher information to weigh the importance of weights before performing SVD-based low-rank approximation. In this work, we focus on reducing the KV cache's memory footprint by storing low-dimensional keys and values to efficiently enable long sequence lengths for LLMs.

\section{Methodology}
This section describes Eigen Attention, which achieves KV cache compression by performing the attention operation in a low-rank space determined by a few basis vectors. We first discuss generating these principal basis vectors and then describe our proposed approach detailed in Algorithm \ref{alg:eigen_attn}. 

\subsection{Basis Vector Generation}\label{sec:basis_generation}
The first step towards computing attention in low-rank space is determining the principle basis vectors that span this low-dimensional space. To achieve this, we create representation matrices $\mathbf{R}^Q_{l,i}$, $\mathbf{R}^K_{l,i}$ and $\mathbf{R}^{V}_{l,i}$ corresponding to query, key, and value respectively, for each layer $l$ and attention head $i$. We drop the layer index $l$ for ease of clarity. These representation matrices are obtained by performing a forward pass for $n_s$ samples taken from the calibration dataset, as shown in Equation \ref{eq:repres}. Note that this calibration dataset is a subset of WikiText \cite{wikitext-arxiv2016}, which is commonly used for pretraining LLMs. 
\begin{equation}
\begin{split}\label{eq:repres}
    \mathbf{R}^{Q}_{i} &= [(\mathbf{Q}^1_{i})^T, (\mathbf{Q}^2_{i})^T, ... , (\mathbf{Q}^{n_s}_{i})^T] \\
    \mathbf{R}^{K}_{i} &= [(\mathbf{K}^1_{i})^T, (\mathbf{K}^2_{i})^T, ... , (\mathbf{K}^{n_s}_{i})^T] \\
    \mathbf{R}^{V}_{i} &= [(\mathbf{V}^1_{i})^T, (\mathbf{V}^2_{i})^T, ... , (\mathbf{V}^{n_s}_{i})^T] \\
\end{split}
\end{equation}
Here, $\mathbf{R}^{Q/K/V}_{i}\in \mathbb{R}^{(n_s.n)\times d_h}$. The key and query representation matrices are concatenated as $\mathbf{R}^{KQ}_{i} = [\mathbf{R}^{Q}_{i}, \mathbf{R}^{K}_{i}]$ followed by an SVD operation $\mathbf{R}^{KQ}_{i}$ $\xrightarrow[]{\text{SVD}}$  $\Sigma^{d_h}_{i=1} \sigma_i u_i v^T_i$. We obtain a $r$-rank approximation  $(\mathbf{R}^{KQ}_{i})_r$ according to the following criteria~\cite{gpm},
\begin{equation}\label{eq:svd}
    ||(\mathbf{R}^{KQ}_{i})_r||^2_F \geq \epsilon_{th}||\mathbf{R}^{KQ}_{i}||^2_F
\end{equation}
$\epsilon_{th}$ is the threshold hyperparameter determining the degree of low-rank approximation. We create a unified basis for key and query to aid in low-rank attention, as shown in Section \ref{sec:eigenattention}. The orthogonal basis vectors spanning the low-rank space for key and query are given by $\mathbf{U}^{KQ}_{i} = [u_1, u_2, ..., u_r]$, which we represent concisely as $\mathbf{U}^{K}_{i}$. Similarly, we generate $\mathbf{U}^{V}_{i}$ for the value representation matrix $\mathbf{R}^{V}_{i}$. Please refer to Appendix \ref{apex:svd} for more details on SVD. Note that a unique low-rank basis is obtained for each head in the attention layer, and we keep the rank $r$ the same across heads by taking the maximum rank across heads. Figure \ref{fig:low_rankness} (a), (b) show the spectrum of eigenvalue distribution of the keys and values for OPT-30b, with keys having a lower rank than the values. In Figure \ref{fig:low_rankness} (c), (d), we plot rank $r$ obtained by keeping $\epsilon_{th}$ as 0.9 for different layers of the model. As shown, some layers' dimensions can be reduced to nearly zero.

\subsection{Eigen Attention}\label{sec:eigenattention}
To understand our proposed low-rank attention, consider the basis vectors $\mathbf{U}^{K}_{i}, \mathbf{U}^V_i$ such that $(\mathbf{U}_i^{K/V})^T.\mathbf{U}^{K/V}_{i}$= $\mathbf{I}$ due to orthogonality. The attention inputs can be projected to the low-rank space spanned by these vectors as,
\begin{equation}
\begin{split}
    \mathbf{Q}'_i &= \mathbf{Q}_i \mathbf{U}^{K}_{i} (\mathbf{U}_i^{K})^T\\
    \mathbf{K}'_i &= \mathbf{K}_i \mathbf{U}^{K}_{i} (\mathbf{U}_i^{K})^T\\
    \mathbf{V}'_i &= \mathbf{V}_i \mathbf{U}^{V}_{i} (\mathbf{U}^{V}_{i})^T
\end{split}
\end{equation}

Here, $\{\mathbf{Q}'_i, \mathbf{K}'_i, \mathbf{V}'_i\} \in \mathbb{R}^{n\times d_h}$ are low rank attention inputs with rank $r$. Then, we compute Eigen Attention as follows: 
\begin{equation}
\begin{split}
    \mathbf{A}'&=\text{Softmax}(\frac{\mathbf{Q}'_i \mathbf{K}_i^{'T}}{\sqrt{d_h}})\mathbf{V}'_i \\
    &=\text{Softmax}(\frac{\mathbf{Q}_i\mathbf{U}^{K}_{i} (\mathbf{U}_i^{K})^T \mathbf{{K}^T}}{\sqrt{d_h}})\mathbf{V}_i\mathbf{U}^{V}_{i} (\mathbf{U}^{V}_{i})^T
\end{split}
\end{equation}

For an appropriate rank $r$, $\mathbf{U}^{K/V}_{i} (\mathbf{U}_i^{K/V})^T \approx \mathbf{I}$ \cite{Yu_Wu_2023}. Hence, attention with low-rank inputs (i.e., Eigen Attention) can approximate the full rank attention ($\mathbf{A}' \approx \mathbf{A}$). Further, the basis vectors $\mathbf{U}^{K}_i$ and $\mathbf{U}^{V}_i$ used to compute Eigen Attention can be seamlessly merged with the weight projection matrices of the attention layer:
\begin{equation}\label{eq:merge_to_weights}
\begin{split}
    \mathbf{W}^{Q'}_i &\leftarrow \mathbf{W}^{Q}_i \mathbf{U}^{K}_{i} \\
    \mathbf{W}^{K'}_i &\leftarrow \mathbf{W}^{K}_i \mathbf{U}^{K}_{i} \\
    \mathbf{W}^{V'}_i &\leftarrow \mathbf{W}^{V}_i \mathbf{U}^{V}_{i} \\
    \mathbf{W}^{O'}_i &\leftarrow (\mathbf{U}^{V}_{i})^T \mathbf{W}^{O}_i
\end{split}
\end{equation}
Here, $\{\mathbf{W}^{Q'}_i, \mathbf{W}^{K'}_i, \mathbf{W}^{V'}_i\}$ $\in$ $\mathbb{R}^{d_h\times r}$ and $\mathbf{W}^{O'}_i \in \mathbb{R}^{r\times d_h}$. This transformation reduces the output dimension of projection matrices, effectively decreasing the embedding dimension of keys, queries, and values. Consequently, both the number of parameters and floating-point operations (FLOPs) in the attention layer are reduced. More importantly, Eigen Attention significantly lowers the KV cache memory footprint (refer Table \ref{tab:eigenmetrics}).


\begin{algorithm}[t]
  \caption{Eigen Attention}\label{alg:eigen_attn}
  \begin{algorithmic}[1]
      \Require LLM Decoder layers $\{\mathbf{L}\}^L_{l=1}$, error budget $e_b$, step size $\epsilon_s$ and inputs to first decoder $X_1$.
      \Procedure{EigenAttention()}{}
      \For{$l = 1 ... L$}
        \State $\mathbf{X}_{l+1}, \mathbf{R}^{KQ}, \mathbf{R}^V \gets$ \texttt{forward}$(\mathbf{L}_l, X_l)$ 
        \State $\epsilon^{l}_{th} \gets 1.0$
        \While{$e \leq e_b$}
            \State $\epsilon^{l}_{th} \gets \epsilon^{l}_{th} - \epsilon_s$
            \State $\mathbf{U}^{K}_l \gets$ \texttt{SVD}$(\mathbf{R}^{KQ}_l, \epsilon^{l}_{th})$ 
            \State $\mathbf{U}^V_{l} \gets$ \texttt{SVD}$(\mathbf{R}^{V}_l, \epsilon^{l}_{th})$ \Comment{eq.\ref{eq:svd}}
            \State $\mathbf{W}^Q_l \gets \mathbf{W}^Q_l \mathbf{U}^{K}_l$ 
            \State $\mathbf{W}^K_l \gets \mathbf{W}^Q_l \mathbf{U}^{K}_l$ 
            \State $\mathbf{W}^V_l \gets \mathbf{W}^Q_l \mathbf{U}^{V}_l$ 
            \State $\mathbf{W}^O_l \gets  (\mathbf{U}^{V}_l)^T \mathbf{W}^Q_l$ \Comment{eq. \ref{eq:merge_to_weights}}
            \State $\mathbf{X}'_{l+1} \gets$ \texttt{forward}$(\mathbf{L}_l, X_l)$  
            \State $e_b = \frac{||\mathbf{X}'_{l+1} - \mathbf{X}_{l+1}||^2}{||\mathbf{X}_{l+1}||^2}$ \Comment{ error}
        \EndWhile
    \EndFor
    \State \Return $\{\mathbf{L}\}^L_{l=1}$\Comment{Low-rank Layers}
    \EndProcedure
  \end{algorithmic}
\end{algorithm}

\subsection{Rotational Position Embedding}\label{sec:rope}
Positional information can be incorporated in text processed by LLMs in several ways. Different models employ absolute or relative positional embeddings at different levels of model computations. Recent LLM demonstrations employ rotary positional embeddings (RoPE) \cite{rope-neurocomputing2024}. RoPE transforms the keys and queries before performing the attention operation as shown below:
\begin{equation}
    \mathbf{Q}^{pos}_i = \mathbf{Q}_i\mathbf{R}; \hspace{2mm}\mathbf{K}^{pos}_i = \mathbf{K}_i\mathbf{R}
\end{equation}
LLMs with RoPE are trained with a fixed dimensional $\mathbf{R}$, making them incompatible with any modification to the embedding dimension of the keys or queries. To integrate RoPE with Eigen Attention, we introduce minor modifications. Specifically, we leave the query to be full rank and transform the key back to a high dimension before applying the RoPE rotation matrix. The query, key dot product with Eigen Attention is given by, 
\begin{equation}
    \mathbf{Q}^{pos}_i (\mathbf{K}^{pos}_i)^T =  \mathbf{Q}_i \mathbf{R} \mathbf{R}^T (\mathbf{U}^K)^T (\mathbf{U}^K \mathbf{K}_i) 
\end{equation}
We store the low-dimensional representation of the key (i.e., $\mathbf{U}^K \mathbf{K}_i$)  in the KV cache but perform an additional transformation through $(\mathbf{U}^K)^T$  before applying RoPE (Figure \ref{fig:appendix_rope}). To mitigate the parameter overhead associated with this additional transformation, we propose to share $\mathbf{U}^K$ across all the attention heads. Similar to the standard Eigen Attention, we merge $\mathbf{U}^K$ into $\mathbf{W}^K_i$, with the value computation unchanged.

\begin{table*}[t!]
\caption{Perplexity (PPL, lower is better) results on Wikitext and C4 and Accuracy (Acc, higher is better) results on lm-eval-harness tasks: PiQA, WinoGrande, Arc-e, Arc-c, HellaSwag. Results are evaluated at different levels of KV cache compression obtained by Eigen Attention. The baseline represents standard attention with an uncompressed KV cache. The performance degradation from the baseline is shown in brackets.} \label{tab:main_results}
\centering
\begin{adjustbox}{width=1\textwidth}
\small
\begin{tabular}{@{\extracolsep{6pt}}ccccccccccc}
\hline
\multirow{2}{*}{Model}    & \multirow{2}{*}{Params} & \multirow{2}{*}{KV Cache} & \multicolumn{2}{c}{PPL $\downarrow$}      & \multicolumn{6}{c}{Acc $\uparrow$}                                      \\ \cline{4-5} \cline{6-11}
                          &                             &                           & Wikitext      & C4            & PiQA & WinoG & Arc-e & Arc-c & HellaS & Avg-Acc          \\ \hline
\multirow{8}{*}{OPT}      & \multirow{4}{*}{30b}        & Baseline                  & 9.56          & 10.69         & 0.78 & 0.68       & 0.70  & 0.35  & 0.54      & 0.61         \\  
                          &                             & 0.8x                      & 9.61 (+0.05)  & 10.75 (+0.06) & 0.78 & 0.67       & 0.70  & 0.34  & 0.54      & 0.61 (-0.00)  \\
                          &                             & 0.7x                      & 9.94 (+0.38)  & 10.98 (+0.29) & 0.78 & 0.66       & 0.69  & 0.35  & 0.54      & 0.60 (-0.01) \\
                          &                             & 0.6x                      & 10.80 (+1.24) & 11.47 (+0.78) & 0.76 & 0.64       & 0.69  & 0.32  & 0.52      & 0.59 (-0.02) \\ \cline{2-11} 
                          & \multirow{4}{*}{66b}        & Baseline                  & 9.34          & 10.28         & 0.79 & 0.69       & 0.72  & 0.37  & 0.56      & 0.63         \\ 
                          &                             & 0.8x                      & 9.36 (+0.02)  & 10.29 (+0.01) & 0.79 & 0.69       & 0.72  & 0.37  & 0.56      & 0.63 (-0.00)  \\
                          &                             & 0.7x                      & 9.51 (+0.17)  & 10.40 (+0.12) & 0.78 & 0.68       & 0.72  & 0.37  & 0.56      & 0.62 (-0.01) \\
                          &                             & 0.6x                      & 9.68 (+0.34)  & 10.54 (+0.26) & 0.78 & 0.68       & 0.70  & 0.36  & 0.55      & 0.62 (-0.01) \\ \hline
\multirow{8}{*}{MPT}      & \multirow{4}{*}{7b}         & Baseline                  & 7.68          & 9.60          & 0.79 & 0.69       & 0.75  & 0.41  & 0.57      & 0.64         \\  
                          &                             & 0.8x                      & 8.02 (+0.34)  & 10.14 (+0.54) & 0.79 & 0.68       & 0.74  & 0.39  & 0.56      & 0.63 (-0.01) \\
                          &                             & 0.7x                      & 8.45 (+0.77)  & 10.80 (+1.20)  & 0.78 & 0.68       & 0.72  & 0.38  & 0.54      & 0.62 (-0.02) \\
                          &                             & 0.6x                      & 9.61 (+1.93)  & 12.26 (+2.66) & 0.76 & 0.68       & 0.69  & 0.35  & 0.52      & 0.60 (-0.04) \\ \cline{2-11} 
                          & \multirow{4}{*}{30b}        & Baseline                  & 6.40          & 8.44          & 0.80 & 0.70       & 0.79  & 0.47  & 0.60      & 0.67         \\ 
                          &                             & 0.8x                      & 6.48 (+0.08)  & 8.55 (+0.11)  & 0.81 & 0.71       & 0.79  & 0.47  & 0.60      & 0.68 (+0.01) \\
                          &                             & 0.7x                      & 6.66 (+0.26)  & 8.82 (+0.38)  & 0.80 & 0.71       & 0.78  & 0.46  & 0.59      & 0.67 (-0.00)  \\
                          &                             & 0.6x                      & 7.01 (+0.61)  & 9.38 (+0.94)  & 0.79 & 0.70       & 0.77  & 0.44  & 0.57      & 0.66 (-0.01) \\ \hline
\multirow{12}{*}{Llama-2} & \multirow{4}{*}{7b}         & Baseline                  & 5.47          & 6.97          & 0.78 & 0.69       & 0.76  & 0.43  & 0.57      & 0.65         \\ 
                          &                             & 0.8x                      & 5.96 (+0.49)  & 7.82 (+0.85)  & 0.76 & 0.66       & 0.72  & 0.40  & 0.54      & 0.61 (-0.04) \\
                          &                             & 0.7x                      & 6.28 (+0.81)  & 8.55 (+1.58)  & 0.75 & 0.63       & 0.70  & 0.38  & 0.52      & 0.60 (-0.05) \\
                          &                             & 0.6x                      & 7.48 (+2.01)  & 10.07 (+3.10)  & 0.74 & 0.63       & 0.65  & 0.33  & 0.48      & 0.56 (-0.09) \\ \cline{2-11} 
                          & \multirow{4}{*}{13b}        & Baseline                  & 4.88          & 6.47          & 0.79 & 0.72       & 0.79  & 0.48  & 0.60      & 0.68         \\  
                          &                             & 0.8x                      & 5.06 (+0.18)  & 6.77 (+0.30)   & 0.78 & 0.72       & 0.78  & 0.45  & 0.59      & 0.66 (-0.02) \\
                          &                             & 0.7x                      & 5.32 (+0.44)  & 7.19 (+0.72)  & 0.77 & 0.70       & 0.77  & 0.45  & 0.57      & 0.65 (-0.03) \\
                          &                             & 0.6x                      & 6.10 (+1.22)  & 9.34 (+2.87)  & 0.73 & 0.65       & 0.70  & 0.36  & 0.48      & 0.58 (-0.10)  \\ \cline{2-11} 
                          & \multirow{4}{*}{70b}        & Baseline                  & 3.32          & 5.52          & 0.82 & 0.78       & 0.83  & 0.54  & 0.65      & 0.72         \\  
                          &                             & 0.8x                      & 3.44 (+0.12)  & 5.74 (+0.22)  & 0.81 & 0.76       & 0.81  & 0.51  & 0.63      & 0.71 (-0.01) \\
                          &                             & 0.7x                      & 3.60 (+0.28)  & 5.95 (+0.43)  & 0.81 & 0.76       & 0.81  & 0.51  & 0.61      & 0.70 (-0.02) \\
                          &                             & 0.6x                      & 3.78 (+0.46)  & 6.23 (+0.71)  & 0.80 & 0.75       & 0.81  & 0.50  & 0.59      & 0.69 (-0.03) \\ \hline
\multirow{8}{*}{Llama-3}  & \multirow{4}{*}{8b}         & Baseline                  & 6.14          & 8.88          & 0.80 & 0.73       & 0.80  & 0.51  & 0.60      & 0.69         \\ 
                          &                             & 0.8x                      & 6.72 (+0.58)  & 9.93 (+1.05)  & 0.78 & 0.73       & 0.78  & 0.48  & 0.57      & 0.67 (-0.02) \\
                          &                             & 0.7x                      & 7.22 (+1.08)  & 10.91 (+2.03) & 0.78 & 0.71       & 0.77  & 0.45  & 0.55      & 0.65 (-0.04) \\
                          &                             & 0.6x                      & 8.51 (+2.37)  & 13.20 (+4.32) & 0.76 & 0.67       & 0.74  & 0.42  & 0.50      & 0.62 (-0.07) \\ \cline{2-11} 
                          & \multirow{4}{*}{70b}        & Baseline                  & 2.85          & 6.73          & 0.82 & 0.81       & 0.87  & 0.60  & 0.66      & 0.75         \\ 
                          &                             & 0.8x                      & 3.22 (+0.37)  & 7.12 (+0.39)  & 0.82 & 0.80       & 0.85  & 0.59  & 0.66      & 0.74 (-0.01) \\
                          &                             & 0.7x                      & 3.59 (+0.74)  & 7.48 (+0.75)  & 0.82 & 0.79       & 0.84  & 0.57  & 0.65      & 0.73 (-0.02) \\
                          &                             & 0.6x                      & 5.54 (+2.69)  & 10.39 (+3.66) & 0.78 & 0.73       & 0.77  & 0.45  & 0.54      & 0.65 (-0.10)  \\ \hline
\end{tabular}
\end{adjustbox}
\end{table*}

\subsection{Layer-wise Rank Allotment}
Eigen Attention introduces layer-wise threshold $\epsilon_{th}$, which determines the accuracy of low-rank approximation according to Equation \ref{eq:svd}. We observe that the same $\epsilon_{th}$ across attention layers introduces different errors at the output of the LLM decoder layer. This implies that the layers that incur lower errors due to the low-rank approximation can be further compressed by lowering the $\epsilon_{th}$. To achieve this, we introduce a layer-wise threshold selection methodology based on the normalized output error of each decoder layer. The threshold $\epsilon_{th}$ is reduced by a step size $\epsilon_s$ until the decoder layer output reaches a specified layerwise error budget. This condenses the layer-wise rank search to two hyperparameters (i.e., error budget $e_b$, and step size $\epsilon_s$), which is kept the same for all LLM layers. Error budget $e_b$ can be increased to increase KV cache compression (Table \ref{tab:appendix_errorbudget}). For all our experiments, we keep $\epsilon_s = 0.02$.

\section{Experiments}
\subsection{Setup}
 We evaluate Eigen Attention across three model families: OPT \cite{opt-arxiv2022}, MPT \cite{mpt-blog}, and Llama \cite{llama2-arxiv2023,llama3-blog}, each with distinct position encoding schemes. OPT employs learnable position embeddings, MPT utilizes AliBi \cite{alibi-arxiv2021}, and the Llama model family employs RoPE \cite{rope-neurocomputing2024}. We conduct evaluations on both language generation and zero-shot tasks. The language generation tasks include perplexity evaluation on Wikitext-2 \cite{wikitext-arxiv2016} and C4 \cite{c4-arxiv2021} datasets. The zero-shot tasks are obtained from lm-eval-harness framework \cite{evalharness-github2023}: PiQA \cite{piqa-aaai2020}, Winogrande (WinoG) \cite{winogrande-acm2021}, Arc-easy/challenge \cite{arc-arxiv2018}, and HellaSwag (HellaS) \cite{hellaswag-arxiv2019}. 

To emphasize that our approach is orthogonal to existing compression techniques, we implement it alongside Grouped Query Attention \cite{groupedqueryattention-arxiv2023} (present in Llama-2 70b and Llama-3) and Quantization \cite{kivi-arxiv2024}.

\begin{table*}[ht!]
\caption{Perplexity and Accuracy results after fine-tuning. The baseline represents standard attention with an uncompressed KV cache. Degradation from baseline is shown in brackets.}\label{tab:finetuning_eigenattn}
\centering
\begin{adjustbox}{width=1\textwidth}
\small
\begin{tabular}{@{\extracolsep{6pt}}cccccccccc}
\hline
\multirow{2}{*}{Model}       & \multirow{2}{*}{KV Cache} & \multicolumn{2}{c}{PPL $\downarrow$}     & \multicolumn{6}{c}{Acc $\uparrow$}                                      \\ \cline{3-4} \cline{5-10} 
                             &                           & Wikitext     & C4            & PiQA & WinoG & Arc-e & Arc-c & HellaS & Avg-Acc          \\ \hline
\multirow{2}{*}{MPT-7b}      & Baseline                  & 7.68         & 9.6           & 0.79 & 0.69       & 0.75  & 0.41  & 0.57      & 0.64         \\
                             & 0.6x                      & 9.21 (+1.53) & 11.44 (+2.14) & 0.79 & 0.69       & 0.72  & 0.39  & 0.55      & 0.63 (-0.01) \\ \hline
\multirow{2}{*}{Llama-2-7b}  & Baseline                  & 5.47         & 6.97          & 0.78 & 0.69       & 0.76  & 0.43  & 0.57      & 0.65         \\
                             & 0.6x                      & 6.55 (+1.07) & 8.55 (+1.58)  & 0.78 & 0.67       & 0.74  & 0.41  & 0.54      & 0.63 (-0.02) \\ \hline
\multirow{2}{*}{Llama-2-13b} & Baseline                  & 4.88         & 6.47          & 0.79 & 0.72       & 0.79  & 0.48  & 0.60      & 0.68         \\
                             & 0.6x                      & 5.64 (+0.76) & 7.77 (+1.30)  & 0.78 & 0.68       & 0.75  & 0.44  & 0.58      & 0.65 (-0.03) \\ \hline
\multirow{2}{*}{Llama-3-8b}  & Baseline                  & 6.14         & 8.88          & 0.80 & 0.73       & 0.80  & 0.51  & 0.60      & 0.69         \\
                             & 0.6x                      & 7.60 (+1.47) & 11.44 (+2.56) & 0.80 & 0.70       & 0.79  & 0.48  & 0.58      & 0.67 (-0.02) \\ \hline
\end{tabular}
\end{adjustbox}
\end{table*}
\begin{figure}[t!]
\includegraphics[width=0.9\linewidth]{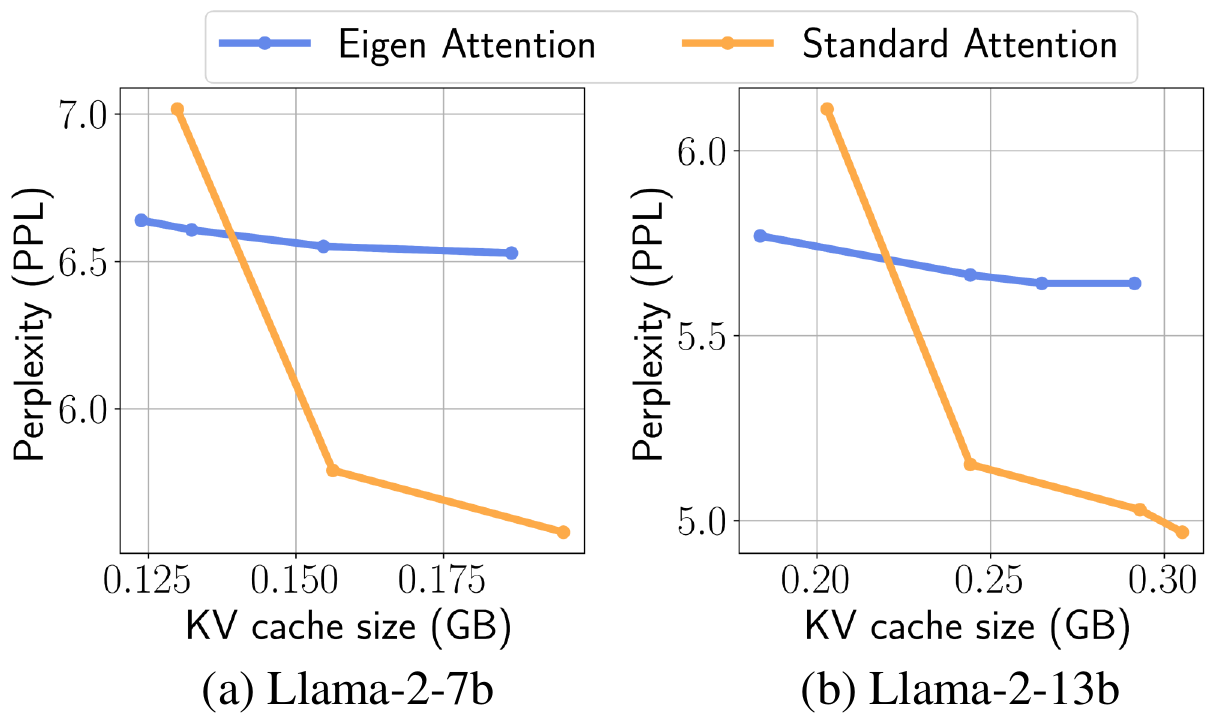}
\caption{PPL on Wikitext with different KV cache sizes in GB ($n$ = 2048) obtained via different quantization precision and group size. For Eigen Attention, we compress the KV cache to 0.6x and then apply quantization.}\label{fig:quantization}
\end{figure}

\subsection{Results}\label{sec:results}
We demonstrate the compression benefits achieved by Eigen Attention through our results in Table \ref{tab:main_results} on various families and sizes of models for a number of language tasks. In particular, we report perplexity (PPL) on Wikitext and C4 datasets and accuracy (Acc) on various zero-shot benchmarks at three KV cache compression ratios: 0.8x, 0.7x, and 0.6x. As expected, increasing the degree of KV cache compression increases the perplexity while reducing accuracy on benchmark tasks. On average, perplexity increases by 0.32, 0.69, and 1.79 while accuracy drops by 1\%, 2\%, and 3\% at 0.8x, 0.7x, and 0.6x KV cache compression, respectively. Within a model family, we find larger models to be more resilient to KV cache compression. Notably, the OPT-66b model incurs only 1\% degradation in zero-shot accuracy with a 40\% compression for the KV cache. Within the Llama-2 family, the average PPL gap to baseline reduces from 2.56 in the 7b model to 0.59 in the 70b parameter model. Across these three distinct model families, we find the KV cache of MPT models to be the most compressible while the Llama-3 models to be the least. For an iso-parameter size of 70b, the Llama-2 model is more resilient to KV cache compression than the Llama-3 model, with both employing grouped query attention. 

\textbf{Eigen Attention followed by Fine-tuning:} We attempt to mitigate the increase in perplexity and the degradation in accuracy observed with the smaller LLMs by fine-tuning. We use LoRA \cite{lora-arxiv2021} to fine-tune these models on the Alpaca dataset \cite{alpaca-github2023} for 2000 steps. Specifically, we fine-tune the query, key, value, and output projection matrices in the attention layer for MPT, Llama-2, and Llama-3 models and report the results in Table \ref{tab:finetuning_eigenattn}. Fine-tuning helps improve the performance of Eigen Attention models, making them perform closer to the baseline. The perplexity gap to baseline reduces from 2.56 (before fine-tuning) to 1.55 after fine-tuning. Similarly, the average zero-shot accuracy gap is reduced from 7\% (before fine-tuning) to within 2\% of baseline.
We find the most improvements in Llama-3-8b, while the least improvements with MPT-7b LLM. 
\begin{table}[]
\caption{Comparison of Eigen Attention with H$_2$O. Combination of Eigen Attention and H$_2$O leads to the best tradeoff between accuracy and KV cache size.}\label{tab:h2o}
\centering
\begin{adjustbox}{width=1\linewidth}
\small
\begin{tabular}{cccc}
\hline
\multirow{2}{*}{Model}      & \multirow{2}{*}{Method} & \multirow{2}{*}{KV Cache} & \multirow{2}{*}{Avg. 0-shot Acc $\uparrow$} \\
                            &                         &                           &                                  \\ \hline
\multirow{4}{*}{Llama-2-7b} & H$_2$O                   & 0.2x                       & 0.54                             \\
                            & H$_2$O                   & 0.4x                       & 0.64                             \\
                            & Eigen Attn              & 0.6x                       & 0.63                             \\
                            & Eigen Attn + H$_2$O      & 0.24x                      & 0.62                             \\ \hline
\multirow{4}{*}{Llama-3-8b} & H$_2$O                   & 0.2x                       & 0.56                             \\
                            & H$_2$O                   & 0.4x                       & 0.67                             \\
                            & Eigen Attn              & 0.6x                       & 0.67                             \\
                            & Eigen Attn + H$_2$O      & 0.24x                      & 0.65                             \\ \hline
\end{tabular}
\end{adjustbox}
\end{table}
\begin{table}[t!]
\caption{Average latency per attention layer during token generation phase.}
\label{tab:gpu_latency}
\centering
\begin{adjustbox}{width=0.85\linewidth}
\small
\begin{tabular}{ccc}
\hline
Model       & Baseline Attn & Eigen Attn 0.6x \\ \hline
OPT-66b     & 1.57 ms  & 0.62 ms (-60\%) \\
Llama-2-70b & 0.79 ms  & 0.77 ms (-3\%)  \\
Llama-3-70b & 0.74 ms  & 0.80 ms (+8\%)  \\ \hline
\end{tabular}
\end{adjustbox}
\end{table}

\textbf{Eigen Attention with Quantization:} 
We compare performance after quantizing key and value in standard and Eigen Attention. Note that the KV cache in Eigen Attention is first compressed to 0.6x before applying quantization.
Figure \ref{fig:quantization} shows perplexity for the Llama-2-7b and Llama-2-13b models on the Wikitext \cite{wikitext-arxiv2016} dataset across various KV cache sizes obtained by performing different levels of quantization. Similar to KIVI \cite{kivi-arxiv2024} and KV-Quant \cite{kvquant-arxiv2024}, we perform per channel quantization for key and per token quantization for value. We implement integer quantization at different precision and group sizes, leading to varied KV cache sizes (Table \ref{tab:appendix_quantization}). For iso-KV cache size, the key and value are quantized to lower precision in quantized standard attention as compared to quantized Eigen Attention. We make two observations: (1) at large KV cache size, the quantized standard attention outperforms quantized Eigen Attention because less error is incurred due to quantization compared to the low-rank decomposition of attention matrices, (2) as quantization precision is reduced to lower the KV cache size, quantized Eigen Attention outperforms quantized standard attention. This is due to a much more severe quantization in standard attention, which induces higher approximation error than Eigen Attention. 

\textbf{Eigen Attention with token pruning:} We compare Eigen Attention with a popular token pruning approach H$_2$O \cite{h20-neurips2023}, which only keeps K and V corresponding to a subset of tokens within the KV Cache.  Table \ref{tab:h2o} shows the average zero shot accuracy comparison at different KV cache compression ratios. We observe that H$_2$O at 0.4x KV cache achieves similar accuracy to Eigen Attention at 0.6x KV cache. Superior performance of H$_2$O over Eigen Attention can be attributed to the dynamic nature of the algorithm. While Eigen Attention compresses KV cache based on statistics derived offline, H$_2$O is able to achieve better generalization by making decisions to eject tokens from KV cache at runtime. However, both Eigen Attention and H$_2$O compress KV cache along different dimensions and can be combined to achieve further compression as seen in Table \ref{tab:h2o}. The combination of Eigen Attention and H$_2$O achieves the best performance highlighting the orthogonality of both approaches. By reducing the hidden dimension of K and V by 60\% using Eigen Attention and using H$_2$O to keep only 40\% of tokens in KV cache, we are able to compress KV cache to 0.24x while achieving accuracy comparable to Eigen Attention at 0.6x and H$_2$O at 0.4x KV cache. 
\begin{table}[t!]
\caption{Total inference latency (in seconds) for Llama-3-8b model at different prompt and generation length.}
\label{tab:gpu_latency_2}
\centering
\begin{adjustbox}{width=\linewidth}
\small
\begin{tabular}{ccccc}
\hline
Batch Size & \begin{tabular}[c]{@{}c@{}}Prompt \\ tokens\end{tabular} & \begin{tabular}[c]{@{}c@{}}Generated \\ tokens\end{tabular} & \begin{tabular}[c]{@{}c@{}}Baseline\\ Attn\end{tabular} & \begin{tabular}[c]{@{}c@{}}Eigen Attn\\ (0.6x)\end{tabular} \\ \hline
1          & 8192                                                     & 2048                                                        & 76                                                      & 82 (+8\%)                                                   \\
1          & 8192                                                     & 16384                                                       & 785                                                     & 880 (+12\%)                                                 \\
1          & 8192                                                     & 32768                                                       & 2048                                                    & 2350 (+13\%)                                                \\
\hline
32         & 2048                                                     & 512                                                         & 180                                                     & 156 (-13\%)                                                 \\
32         & 2048                                                     & 2048                                                        & 826                                                     & 859 (+4\%)                                                  \\
32         & 2048                                                     & 4096                                                        & \textbf{OOM}                                            & 2366                                                        \\ \hline
\end{tabular}
\end{adjustbox}
\end{table}

\textbf{Latency Comparisons:} We compare the latency of Eigen Attention and standard attention baseline for different models. In Table \ref{tab:gpu_latency}, we show average latency per attention layer during the token generation phase for generating 128 tokens with a 2048-token synthetic prompt on 2 NVIDIA A100 GPUs. We observe an impressive 60\% latency improvement for the OPT-66b model, which can be attributed to the FLOPs reduction from Eigen Attention and the reduced latency in fetching the compressed KV cache from memory. For models with RoPE embedding (i.e., Llama-2-70b and Llama-3), we perform an additional transformation before the attention dot product (Figure \ref{fig:appendix_rope}), which diminishes the latency gains achieved with Eigen Attention. For Llama-2-70b, we observe only a 3\% latency improvement, while for Llama-3-70b, there is a latency penalty of 8\%. We further analyze end to end inference latency at different context lengths and batch sizes in Table \ref{tab:gpu_latency_2} for Llama-3-8b on a single NVIDIA A100 GPU. We see that increasing the context length increases the latency overhead with Eigen Attention due to the presence of additional transformation to manage RoPE embedding. For the maximum sequence length of 40k, there is a 13\% latency overhead with Eigen Attention for Llama-3-8b. When batch size is increased, the KV cache size increases which increases the memory bounded nature of inference. In this scenario, additional transformation required by Eigen Attention leads to much lesser latency overhead and often faster inference than baseline attention. Most notably, at high batch size and context length, baseline attention leads to out of memory (OOM) error on GPU while Eigen Attention does not. This highlights the main contribution of our work: to enable long context inference by compressing KV cache.

\begin{table}[]
\caption{Perplexity and zero-shot accuracy when using different calibration datasets for MPT-7b model.}\label{tab:calibdataset}
\centering
\begin{adjustbox}{width=0.9\linewidth}
\small
\begin{tabular}{ccccc}
\hline
\begin{tabular}[c]{@{}c@{}}Calib.\\ dataset\end{tabular} & KV Cache & \begin{tabular}[c]{@{}c@{}}Wikitext\\ (PPL)\end{tabular} & \begin{tabular}[c]{@{}c@{}}C4\\ (PPL)\end{tabular} & \begin{tabular}[c]{@{}c@{}}Zero-shot  \\ Acc. (\%)\end{tabular} \\ \hline
\multirow{3}{*}{C4}                                      & 0.8      & 8.03                                                     & 9.96                                               & 0.63                                                            \\
                                                         & 0.7      & 8.57                                                     & 10.47                                              & 0.63                                                            \\
                                                         & 0.6      & 9.63                                                     & 11.44                                              & 0.62                                                            \\ \hline
\multirow{3}{*}{Alpaca}                                  & 0.8      & 8.12                                                     & 10.07                                              & 0.64                                                            \\
                                                         & 0.7      & 3.02                                                     & 10.88                                              & 0.62                                                            \\
                                                         & 0.6      & 10.86                                                    & 12.42                                              & 0.61                                                            \\ \hline
\end{tabular}
\end{adjustbox}
\end{table}
\begin{figure}[t!]
\includegraphics[width=0.9\linewidth]{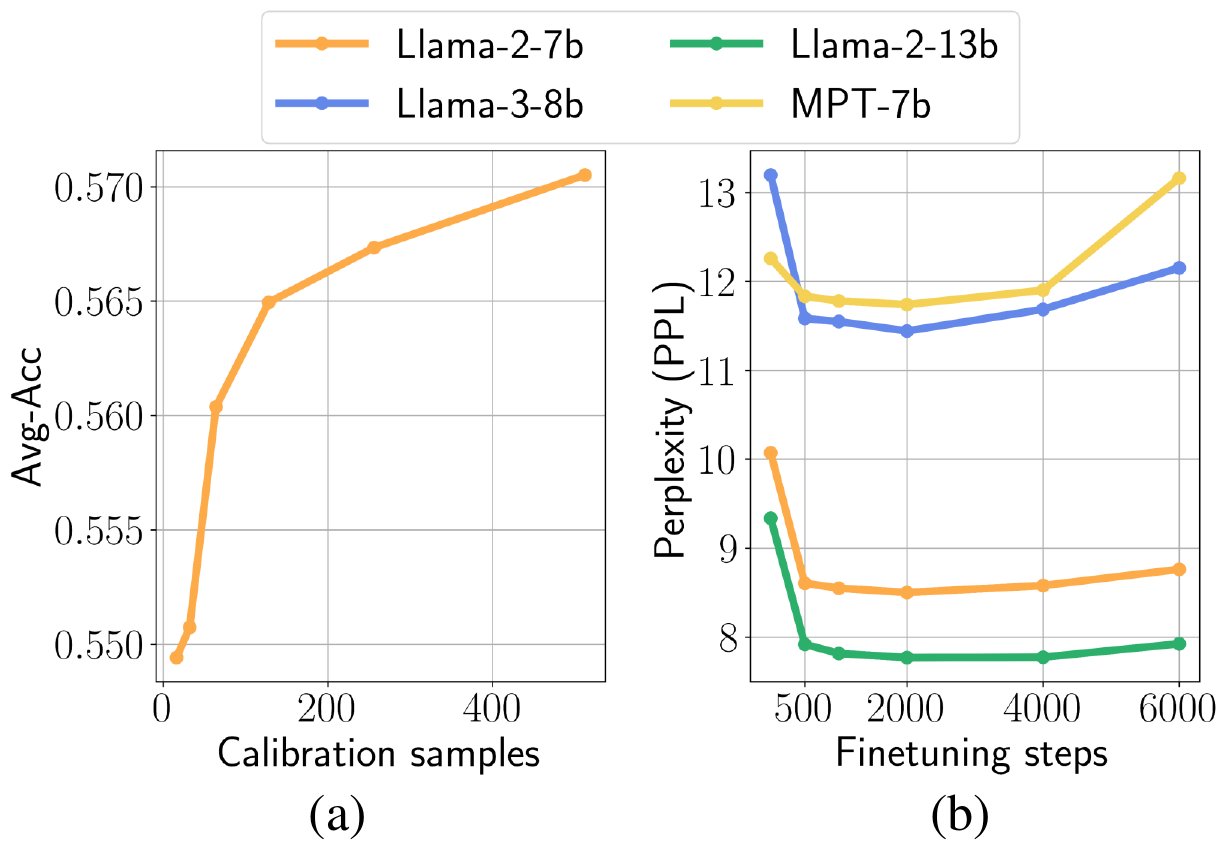}
\vspace{-1.5mm}
\caption{Ablation Study. (a) Average accuracy (Avg-Acc) on zero-shot tasks trained on Llama-2-7b with an increasing number of calibration samples. (b) Perplexity (PPL) vs fine-tuning steps on the C4 dataset for MPT and Llama family of models. }\label{fig:ablation}
\end{figure}

\subsection{Ablation Studies}
We analyze our approach by varying the number of calibration samples used to compute the representation matrix for low-rank approximation and the number of steps used to fine-tune the models.

\textbf{Impact of calibration dataset:} Table \ref{tab:calibdataset} shows the performance of MPT-7b model when different calibration datasets are used for obtaining low rank space of K, V vectors. We select samples from C4 \cite{c4-arxiv2021} and Alpaca \cite{alpaca-github2023} datasets for obtaining the low rank basis. We see that compressing using C4 calibration dataset achieves better perplexity on the C4 dataset itself while achieving slightly lower perplexity on Wikitext dataset (results in Table \ref{tab:main_results}). Performance on zero shot tasks remains similar regardless of the calibration dataset choice. 

\textbf{Number of Calibration Samples:} Figure \ref{fig:ablation}(a) shows average accuracy (Avg-Acc) on zero-shot benchmarks with 0.6x KV cache for different numbers of calibration samples used to generate the representation matrix for low-rank approximation. Increasing the number of calibration samples enhances the Avg-Acc, as more samples lead to a more generalized set of basis vectors. However, more samples also increase the size of representation matrices (Equation \ref{eq:repres}), requiring significantly higher GPU memory for performing SVD. Using more than 128 calibration samples leads to out-of-memory errors. We average representations from samples to handle this, thereby reducing representation matrix dimensions.
For instance, representations from every 2 (4) samples are averaged for 256 (512) samples.

\textbf{Fine-tuning Steps:} In Figure \ref{fig:ablation}(b), we evaluate perplexity (PPL) for the C4 dataset evaluated on the MPT and Llama models for a range of fine-tuning steps. With just 500 steps, the PPL improves by 1.2 on average across all the models. 2000 steps are ideal, beyond which we observe an average increase of 0.61 in the PPL. We posit that this is due to model overfitting on finetuning data.

\subsection{Layerwise Rank Visualization}
\begin{figure}[t!]
    \centering
    \begin{subfigure}{0.24\textwidth}
        \centering
        \includegraphics[height=1.4in]{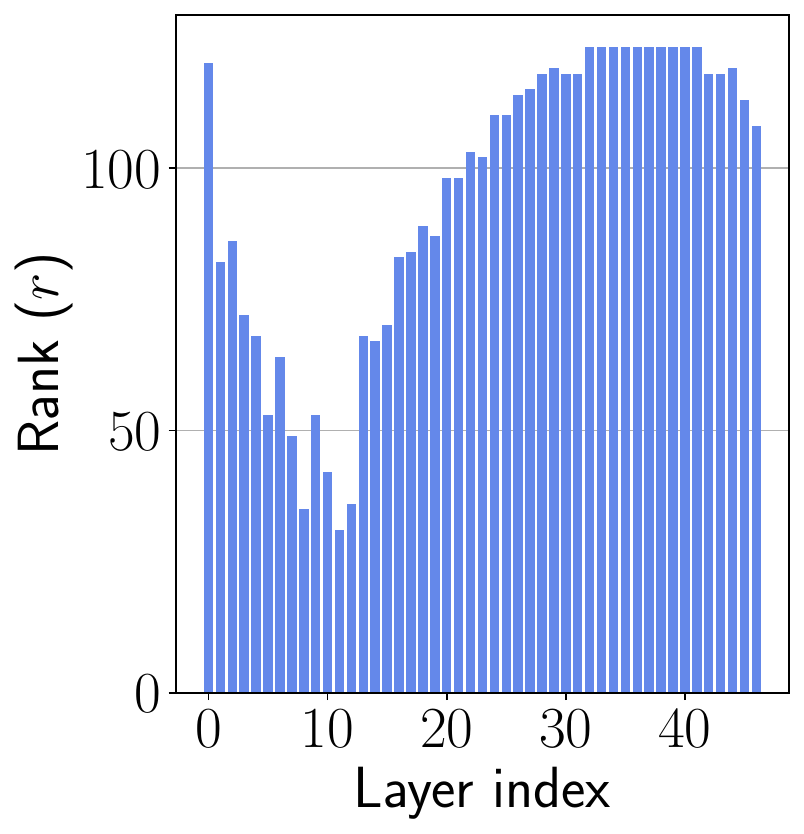}
        \caption{Value}
    \end{subfigure}%
    \begin{subfigure}{0.24\textwidth}
        \centering
        \includegraphics[height=1.4in]{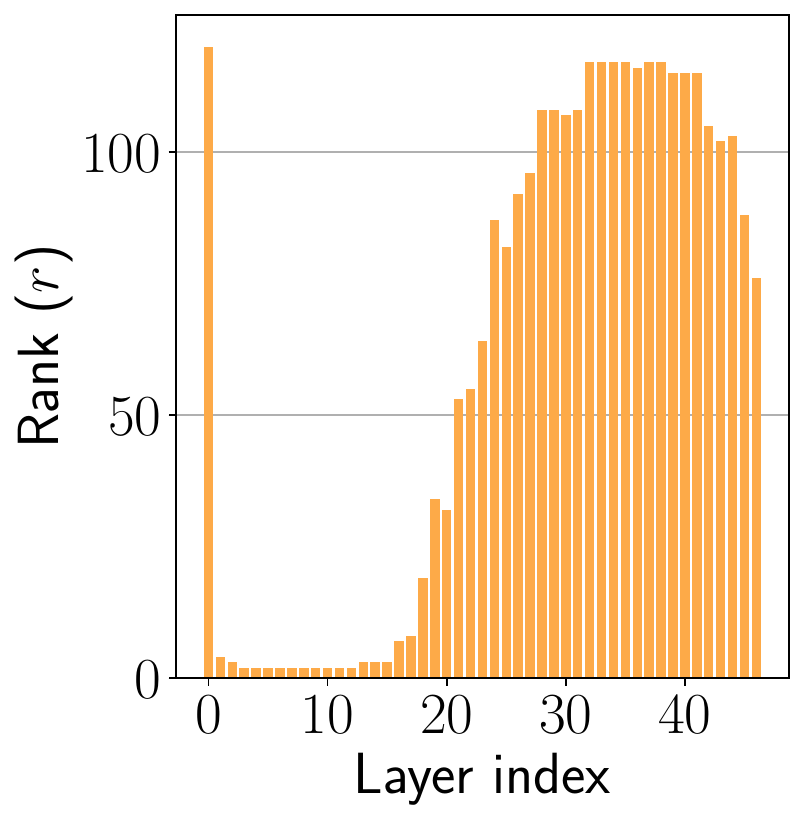}
        \caption{Key}
    \end{subfigure}%
\vspace{-1.5mm}
    \caption{Layerwise rank assignment for key and value determined by Eigen Attention for OPT-30b with 40\% compressed KV cache.}\label{fig:rank_opt30b}
\end{figure}

Figure \ref{fig:rank_opt30b} shows the rank $r$ assigned to key and query projection layers by Eigen Attention at 40\% KV cache compression. We observe that the initial layers (with the exception of first layer) are compressed more than the later layers. Also, keys are assigned a lower rank and are compressed more than values, which concurs with our eigenvalue spectrum analysis in Figure \ref{fig:low_rankness}. Specifically, for 40\% KV cache compression, keys are compressed by 54\% while values are compressed by only 26\%. 

\vspace{-1mm}

\section{Conclusion}
In this work, we propose Eigen Attention, a novel technique that reduces the memory overhead associated with KV cache in LLMs. Eigen Attention is inspired by the observation that keys, queries, and values can be approximated using a few basis vectors, enabling the possibility of performing the attention operation in a low-rank space with minimal performance loss. To achieve this, we project the attention inputs into low-rank subspaces defined by a set of principal basis vectors computed offline using a calibration dataset in a one-shot manner. These projections are integrated into the weight matrices, which are utilized during inference to generate lower dimensional key and value matrices, thereby reducing the KV cache memory footprint. Our approach is orthogonal to existing KV cache compression strategies and can be used in synergy. Extensive experiments across a range of LLMs and language tasks demonstrate that Eigen Attention reduces the KV cache memory footprint by 40\% with minimal performance loss and achieves up to a 60\% improvement in attention operation latency compared to the standard attention baseline. 
\section*{Limitations}
Eigen Attention takes a step towards efficiently enabling longer context lengths, thereby opening avenues for enhancing the capabilities of the current state-of-the-art LLMs. While we demonstrate low-rank basis generation using the Wikitext dataset \cite{wikitext-arxiv2016}, we do not extensively study the best dataset for basis generation. 
Additionally, although our proposed approach has the potential to make LLMs ubiquitous, it does not mitigate the risks of misuse of these models for malicious activities. A strong commitment to user data protection, robust ethical guidelines, and transparency mechanisms is essential to address this issue effectively. 

\section*{Acknowledgements}
This work was supported by the Center for the Co-Design of Cognitive Systems (COCOSYS), a DARPA sponsored JUMP center of Semiconductor Research Corporation (SRC), National Science Foundation and United States Department of Energy.

\bibliography{main_paper}

\appendix
\section{Appendix}
\label{sec:appendix}

\subsection{SVD for Matrix Approximation}\label{apex:svd}
Singular Value Decomposition (SVD) can be used to obtain a low-rank approximation for any matrix $\mathbf{Z} \in \mathbb{R}^{m \times n}$ by factorizing it into three matrices as $\mathbf{Z}= \mathbf{U}\mathbf{\Sigma}\mathbf{V}$. Here, $\mathbf{U} \in \mathbb{R}^{m \times m}$ and $\mathbf{V} \in \mathbb{R}^{n \times n}$ are orthogonal matrices, and $\mathbf{\Sigma}$ is a diagonal matrix which contains the sorted singular values. For $\mathbf{Z}$ with a rank $r\leq \text{min}(m,n)$, it can be expressed as $\mathbf{Z}= \sum_{i=1}^{r} \sigma_i u_i v_i^T$, where $u_i \in \mathbf{U}$, $v_i \in \mathbf{V}$ and $\sigma_i \in \mathbf{\Sigma}$. For a $k$-rank approximation with $k < r$, we have $\mathbf{Z}_k= \sum_{i=1}^{k} \sigma_i u_i v_i^T$ such that the top $k$ values from $\mathbf{\Sigma}$ are chosen to represent $\mathbf{Z}_k$, a low-rank approximation of $\mathbf{Z}$.

\subsection{Eigen Attention with RoPE}
\begin{figure}[h!]
\includegraphics[width=\linewidth]{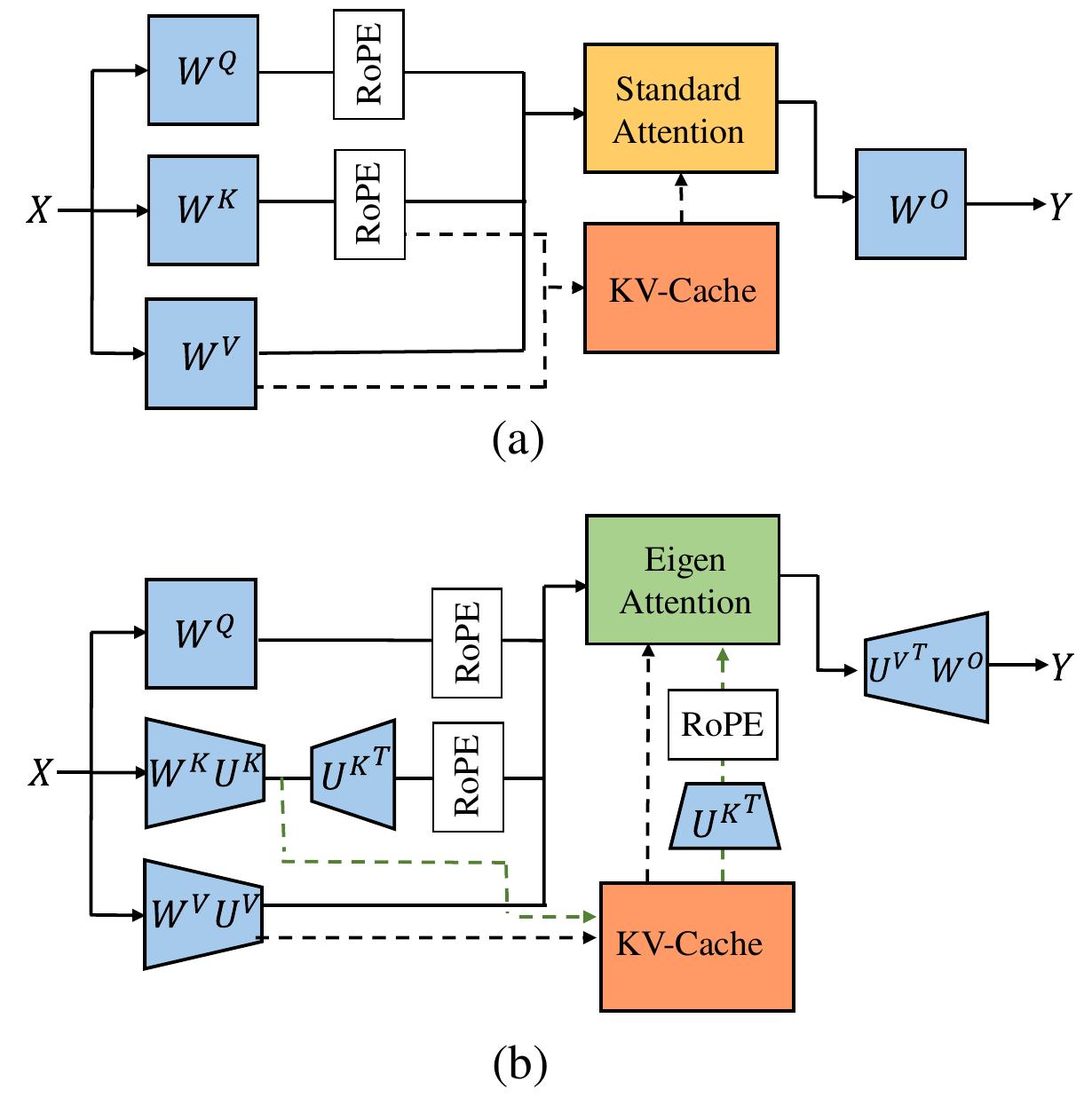}
\caption{Comparison between (a) Standard Attention and (b) Eigen Attention for LLMs with RoPE \cite{llama2-arxiv2023}. Eigen Attention stores low dimensional representation of key and value in KV cache and applies an additional transformation before applying RoPE.}\label{fig:appendix_rope}
\end{figure}

We introduce some modifications to Eigen Attention algorithm to handle compatibility with LLMs employing RoPE empedding (Section \ref{sec:rope}). The comparison of Eigen Attention with standard attention in the presence of RoPE is shown in Figure \ref{fig:appendix_rope}.

\subsection{Quantization Results}
\begin{table}[ht!]
\caption{Perplexity (PPL) on Wikitext after applying quantization to eigen attention and standard attention.}\label{tab:appendix_quantization}
\centering
\begin{adjustbox}{width=1\linewidth}
\small
\begin{tabular}{cccccc}
\hline
Model & Method & Precision & Group Size & KV Cache size (GB) & PPL (Wikitext) \\ \hline
\multirow{8}{*}{Llama-2-13b} & \multirow{4}{*}{Eigen Attention} & 3 & 128 & 0.18 & 5.77 \\
 &  & 4 & 128 & 0.24 & 5.66 \\
 &  & 4 & 32 & 0.26 & 5.64 \\
 &  & 4 & 16 & 0.29 & 5.64 \\ \cline{2-6} 
 & \multirow{4}{*}{Standard Attention} & 2 & 512 & 0.20 & 6.11 \\
 &  & 2 & 32 & 0.24 & 5.15 \\
 &  & 2 & 16 & 0.29 & 5.03 \\
 &  & 3 & 128 & 0.31 & 4.97 \\ \hline
\multirow{7}{*}{Llama-2-7b} & \multirow{4}{*}{Eigen Attention} & 4 & 16 & 0.19 & 6.53 \\
 &  & 4 & 128 & 0.15 & 6.55 \\
 &  & 3 & 32 & 0.13 & 6.61 \\
 &  & 3 & 64 & 0.12 & 6.64 \\ \cline{2-6} 
 & \multirow{3}{*}{Standard Attention} & 2 & 512 & 0.13 & 7.02 \\
 &  & 2 & 32 & 0.16 & 5.79 \\
 &  & 3 & 128 & 0.20 & 5.58 \\ \hline
\end{tabular}
\end{adjustbox}
\end{table}
Section \ref{sec:results} shows results for performing quantization with Eigen Attention and standard attention. The data corresponding to Figure \ref{fig:quantization} is provided in Table \ref{tab:appendix_quantization}.

\subsection{Hyperparameters}
\begin{table}[h!]
\caption{Error budget $e_b$ for Eigen Attention. KV cache size is computed for a sequence length of 2048.}\label{tab:appendix_errorbudget}
\centering
\begin{adjustbox}{width=1\linewidth}
\small
\begin{tabular}{ccccc}
\hline
\multirow{2}{*}{Model}   & \multirow{2}{*}{Parameters} & \multirow{2}{*}{KV cache compression} & \multirow{2}{*}{KV cache size (GB)} & \multirow{2}{*}{$e_b$} \\
                         &                             &                                       &                                     &                               \\ \hline
\multirow{6}{*}{OPT}     & \multirow{3}{*}{30b}        & 0.8x                                  & 2.10                                & 0.008                         \\
                         &                             & 0.7x                                  & 1.87                                & 0.015                         \\
                         &                             & 0.6x                                  & 1.60                                & 0.024                         \\ \cline{2-5} 
                         & \multirow{3}{*}{66b}        & 0.8x                                  & 3.69                                & 0.005                         \\
                         &                             & 0.7x                                  & 3.22                                & 0.012                        \\
                         &                             & 0.6x                                  & 2.80                                & 0.015                         \\ \hline
\multirow{6}{*}{MPT}     & \multirow{3}{*}{7b}         & 0.8x                                  & 0.80                                & 0.011                         \\
                         &                             & 0.7x                                  & 0.70                                & 0.015                         \\
                         &                             & 0.6x                                  & 0.60                                & 0.019                         \\ \cline{2-5} 
                         & \multirow{3}{*}{30b}        & 0.8x                                  & 2.12                                & 0.002                        \\
                         &                             & 0.7x                                  & 1.83                                & 0.003                        \\
                         &                             & 0.6x                                  & 1.59                                & 0.004                        \\ \hline
\multirow{9}{*}{Llama-2} & \multirow{3}{*}{7b}         & 0.8x                                  & 0.83                                & 0.025                         \\
                         &                             & 0.7x                                  & 0.71                                & 0.070                          \\
                         &                             & 0.6x                                  & 0.59                                & 0.090                          \\ \cline{2-5} 
                         & \multirow{3}{*}{13b}        & 0.8x                                  & 1.29                                & 0.035                         \\
                         &                             & 0.7x                                  & 1.12                                & 0.050                          \\
                         &                             & 0.6x                                  & 0.93                                & 0.070                          \\ \cline{2-5} 
                         & \multirow{3}{*}{70b}        & 0.8x                                  & 3.44                                & 0.005                         \\
                         &                             & 0.7x                                  & 3.60                                & 0.010                          \\
                         &                             & 0.6x                                  & 3.78                                & 0.017                         \\ \hline
\multirow{6}{*}{Llama-3} & \multirow{3}{*}{8b}         & 0.8x                                  & 6.72                                & 0.045                         \\
                         &                             & 0.7x                                  & 7.22                                & 0.060                          \\
                         &                             & 0.6x                                  & 8.51                                & 0.090                          \\ \cline{2-5} 
                         & \multirow{3}{*}{70b}        & 0.8x                                  & 3.22                                & 0.025                         \\
                         &                             & 0.7x                                  & 3.59                                & 0.035                         \\
                         &                             & 0.6x                                  & 5.54                                & 0.070                          \\ \hline
\end{tabular}
\end{adjustbox}
\end{table}
We use the Hugging Face Transformers library \cite{huggingface-arxiv2019} to implement Eigen Attention in PyTorch \cite{pytorch-Neurips2019}. All the experiments were performed on NVIDIA A100 (80GB) GPUs. All models are downloaded from  \href{https://huggingface.co/docs/hub/en/index}{Hugging Face Hub}. We show results on: \href{https://huggingface.co/facebook/opt-30b}{OPT-30b}, \href{https://huggingface.co/facebook/opt-66b}{OPT-66b}, \href{https://huggingface.co/mosaicml/mpt-7b}{MPT-7b}, \href{https://huggingface.co/mosaicml/mpt-30b}{MPT-30b}, \href{https://huggingface.co/meta-llama/Llama-2-7b-hf}{Llama-2-7b}, \href{https://huggingface.co/meta-llama/Llama-2-13b-hf}{Llama-2-13b}, \href{https://huggingface.co/meta-llama/Llama-2-70b-hf}{Llama-2-70b},  \href{https://huggingface.co/meta-llama/Meta-Llama-3-8B}{Llama-3-8b} and, \href{https://huggingface.co/meta-llama/Meta-Llama-3-70B}{Llama-3-70b}.

All the models are compressed using 512 ($n_s$) samples of sequence length ($n$) 2048 from Wikitext dataset \cite{wikitext-arxiv2016}. Eigen Attention introduces a hyperparameter for error budget ($e_b$), which is tuned to achieve the required KV cache compression, as listed in Table \ref{tab:appendix_errorbudget}. We keep $\epsilon_s$ = 0.02 for all our runs.

For fine-tuning, we use $lora\_r = 64$, $lora\_alpha = 64$, $sequence\_length = 2048$, $lora\_dropout = 0.05$ and use the default values from the Hugging Face PEFT library \cite{huggingfacepeft-github2022} for all the other hyperparameters. We observe that fine-tuning on the Alpaca dataset \cite{alpaca-github2023} performs better than C4 \cite{c4-arxiv2021}. 

\subsection{Artifact Licenses}
According to their license, all the LLMs used in this paper fall under acceptable use cases. The licenses for the models are linked for perusal: \href{https://huggingface.co/facebook/opt-30b}{OPT-30b}, \href{https://huggingface.co/facebook/opt-66b}{OPT-66b}, \href{https://huggingface.co/datasets/choosealicense/licenses/blob/main/markdown/apache-2.0.md}{MPT-7b}, \href{https://huggingface.co/datasets/choosealicense/licenses/blob/main/markdown/apache-2.0.md}{MPT-30b}, \href{https://huggingface.co/meta-llama/Llama-2-7b-chat-hf/blob/main/LICENSE.txt}{Llama-2-7b}, \href{https://huggingface.co/meta-llama/Llama-2-7b-chat-hf/blob/main/LICENSE.txt}{Llama-2-13b}, \href{https://huggingface.co/meta-llama/Llama-2-7b-chat-hf/blob/main/LICENSE.txt}{Llama-2-70b},  \href{https://huggingface.co/meta-llama/Meta-Llama-3-8B/blob/main/LICENSE}{Llama-3-8b} and, \href{https://huggingface.co/meta-llama/Meta-Llama-3-8B/blob/main/LICENSE}{Llama-3-70b}.

\subsection{Future Work}\label{apex:future}
We describe two key future directions for Eigen Attention: (1) integrating Eigen Attention with efficient LLM serving frameworks like vLLM \cite{vllm-acmsigops2023}, which employ additional approximation techniques (e.g., weight quantization \cite{awq-arxiv2024}) to achieve high throughput inference, and (2) finding the best combination of various compression techniques described in Section \ref{sec:related_works} to achieve extreme KV cache compression.

\end{document}